# Knowledge Discovery of Hydrocyclone's Circuit Based on SONFIS &SORST

H. O. Ghaffari, M. Ejtemaei, M. Irannajad

Amirkabir University of Technology, Faculty of Mining, Metallurgical and Petroleum Engineering, Tehran, Iran

**ABSTRACT:**
This study describes application of some approximate reasoning methods to analysis of hydrocyclone performance. In this manner, using a combining of Self Organizing Map (SOM), Neuro-Fuzzy Inference System (NFIS)-SONFIS- and Rough Set Theory (RST)-SORST-crisp and fuzzy granules are obtained. Balancing of crisp granules and non-crisp granules can be implemented in close-open iteration. Using different criteria and based on granulation level balance point (interval) or a pseudo-balance point is estimated. Validation of the proposed methods, on the data set of the hydrocyclone is rendered.

## 1. INTRODUCTION:

In the design of mineral processing circuit, one of the most important issues is the selection of hydrocyclone in different parts of the circuit. However, prediction of hydrocyclone performance using direct or indirect modeling has its own difficulties. In recent years application of intelligent methods, in data analysis (knowledge discovery) or control process in mineral engineering has been extended (Carvalho and Durao, 2002; Vieira et al., 2005).

The most main distinguished facets of the soft granules are: set theory, interval analysis, fuzzy set, rough set. Each of these theories considers part of uncertainty of information (data, words, pictures...).Due to association of uncertainty and vagueness with the monitored data set, particularly, resulted from the laboratory or industrial tests, accounting relevant approaches such probability, Fuzzy Set Theory (FST) and Rough Set Theory (RST) to knowledge acquisition, extraction of rules and prediction of unknown cases, more than the past have been distinguished.

The RST introduced by Pawlak has often proved to be an excellent mathematical tool for the analysis of a vague description of object (Pawlak, 1991) the adjective vague, referring to the quality of information, means inconsistency, or ambiguity which follows from information granulation. The rough set philosophy is based on the assumption that with every object of the universe, is associated a certain amount of information, expressed by means of some attributes used for

171

object description. The indiscernibility relation (similarity), which is a mathematical basis of the rough set theory, induces a partition of the universe in to blocks of indiscernible objects, called elementary sets, which can be used to build knowledge about a real or abstract world. Precise condition rules can be extracted from a discernibility matrix. Application of RST in different fields of the applied sciences has been reported (Pal et al., 2004).

## 2. EXPERIMENTAL:

Experiments were conducted with the hydrocyclone Test Rig C705 (Figure1a).The verification data is divided into hydrocyclone operations according to the different pressure drop (psi) and solid percent, as the tests run with constant geometrical parameters (diameters of the hydrocyclone=50.8cm, overflow=30m, underflow=7mm).

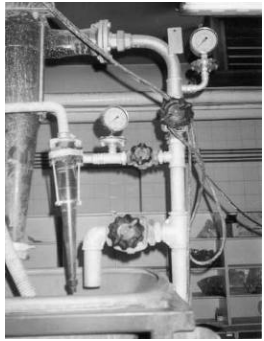
(a)

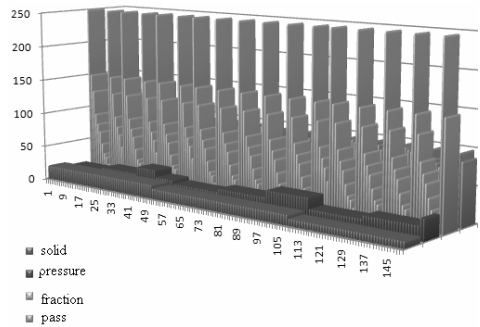
(b)

Figure 1. a) The hydrocyclone that used in this study; b) The overall results of the test on the sample

The sample that used in this study was collected from Qara Ağac kaolin mining of Iran, where the density of our sample is 2.17 g/cm$^3$.

The process has four manipulating variables: Pressure drop (psi), solid percent (%), size fraction (μm) and overflow or underflow state (ascribed 0, 1codes, respectively). The main output of this model is cumulate passing percent (%) that used to control of the split size ($d_{50}$) and, as a direct result, calculation of Imperfection coefficient in the hydrocyclone operations, can be evaluated (Figure1b).

## 3. PROPOSED ALGORITHMS:

In the whole of our algorithms, we use four basic axioms upon the balancing of the successive granules (clusters):

Step (1): dividing the monitored data into groups of training and testing data

Step (2): first granulation (crisp) by SOM or other crisp granulation methods

Step (2-1): selecting the level of granularity randomly or depend on the obtained error from the NFIS or RST (regular neuron growth)

Step (2-2): construction of the granules (crisp).

Step (3): second granulation (fuzzy or rough clusters) by NFIS or RST

Step (3-1): crisp granules as a new data.

Step (3-2): selecting the level of granularity; (Error level, number of rules, strength threshold...)

Step (3-3): checking the suitability. (Close-open iteration: referring to the real data and reinspect closed world)

Step (3-4): construction of fuzzy/rough granules.

Step (4): extraction of knowledge rules

Selection of initial crisp granules can be supposed as "Close World Assumption (CWA)" .But in many applications, the assumption of complete information is not feasible, and only cannot be used. In such cases, an "Open World Assumption (OWA)', where information not known by an agent is assumed to be unknown, is often accepted. Balancing assumption is satisfied by the close-open iterations: this process is a guideline to balancing of crisp and sub fuzzy/rough granules by some random/regular selection of initial granules or other optimal structures and increment of supporting rules (fuzzy partitions or increasing of lower /upper approximations ), gradually.

The overall schematic of Self Organizing Neuro-Fuzzy Inference System - Random and Regular neuron growth-: SONFIS-R has been shown in Figure2.

Determination of granulation level is controlled with three main parameters: range of neuron growth, number of rules and error level. The main benefit of this algorithm is to looking for best structure and rules for two known intelligent system, while in independent situations each of them has some appropriate problems such: finding of spurious patterns for the large data sets, extra-time training of NFIS or SOM.

In second algorithm, apart from employing hard computing methods (hard granules), RST instead of NFIS has been proposed (Figure 3). Applying of SOM as a preprocessing step and scaling tool is second process. Categorization of attributes (inputs/outputs) is transferring of the attribute space to the symbolic appropriate attributes. Because of the generated rules by a rough set are coarse and therefore need to be fine-tuned, here, we have used the preprocessing step on data set to crisp granulation by SOM (close world assumption).

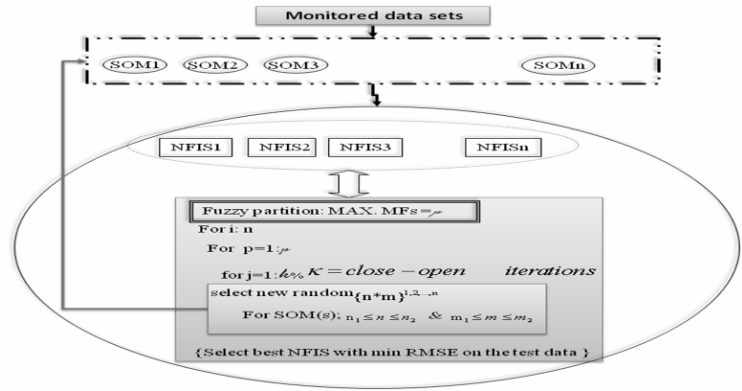

Figure 2. Self Organizing Neuro-Fuzzy Inference System (SONFIS)

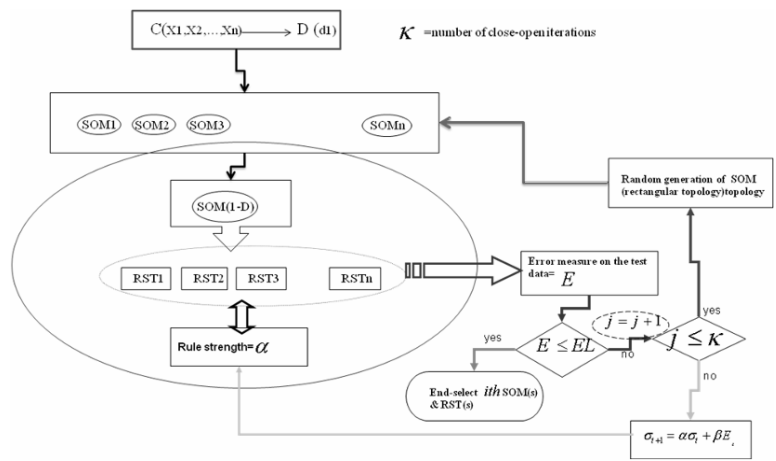

Figure 3. Self Organizing Rough Set Theory-Random neuron growth & adaptive strength factor (SORST-R)

## 4. RESULTS:

Analysis of first situation is started off by setting number of close-open iteration and maximum number of rules equal to 10 and 4 in SONFIS-R, respectively. The

174

error measure criteria in SONFIS are Root Mean Square Error (RMSE), given as below: $RMSE = \sqrt{\dfrac{\sum_{i=1}^{m}(t_i - t_i^*)^2}{m}}$ ;

where $t_i$ is output of SONFIS and $t_i^*$ is real answer; m is the number of test data (test objects). In the rest of paper, let m=19 and number of training data set =150. Figures 3 indicate the results of the aforesaid system. The indicated position in Figure 3a, b states minimum RMSE over the iterations.

It is worth noting that upon this balancing criterion, we may lose the general dominant distribution on the data space. The performance of the obtained fuzzy rules on the test data has been portrayed in Figure 4(a). In employing of second algorithm (Figure3), we use- -for in this case- only exact rules i.e., one decision class in right hand of an if-then rule. Figure 6 depicts the performance of SORST-R over 7 random selection of SOM structure, respectively. The applied Error measure is : $EM = \dfrac{\sum_{i=1}^{m}(d_i^{real} - d_i^{classified})^2}{m}$ ;

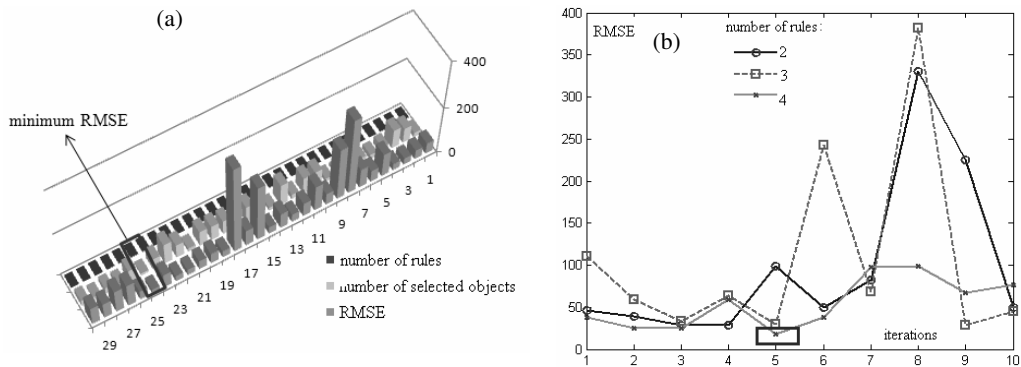

Figure 4. Obtained results by SONFIS-R and the minimum RMSE in 30 iteration - 10 for each rule

175

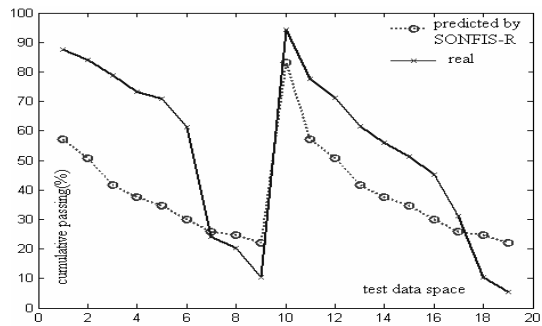

Figure 5. The real and predicted decision on the testing data set with sub-fuzzy granulation

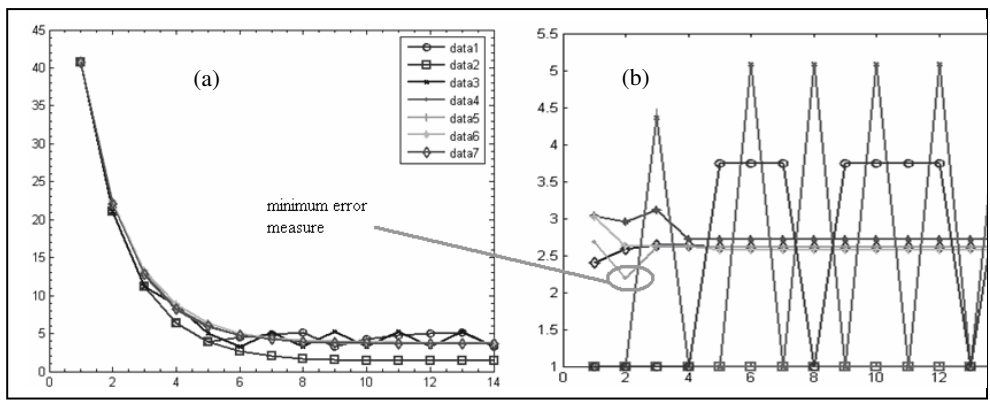

Figure 6. SORST-R results on the Hydrocyclon data set a) strength factor convergence (approximately); b) error measure variations along strength factor updating

In this case, strength factor is adapted with EM, in a linear form. It must be noticed that for unrecognizable objects in test data (elicited by rules) a fix value such 4 is ascribed. So for measure part when any object is not identified, 1 is attributed. This is main reason of such swing of EM in reduced data set 6 (Figure 6-b). Clearly, in data set 5 SORST gains a lowest error (15 neurons in SOM). Total of extracted rules on the training data set (reduced data) were 33.

## 5. CONCLUSION:

This paper has presented two successive granulation methods to knowledge discovery. We have employed our algorithms, to prediction of hydrocyclone performance and extraction of rules which are control the split size. The results show that SONFIS-R with 4 rules gets larger error in continuous scale of data whereas SORST upon 33 rule and with lesser than iteration –on the scaled data set- emerges better outputs.